# Conditional Independence and Markov Properties in Possibility Theory

Jiřina Vejnarová
Laboratory for Intelligent Systems
University of Economics, Prague
and
Institute of Information Theory and Automation
Academy of Sciences of the Czech Republic

## Abstract

Conditional independence and Markov properties are powerful tools allowing expression of multidimensional probability distributions by means of low-dimensional ones. As multidimensional possibilistic models have been studied for several years, the demand for analogous tools in possibility theory seems to be quite natural. This paper is intended to be a promotion of de Cooman's measure-theoretic approach to possibility theory, as this approach allows us to find analogies to many important results obtained in probabilistic framework. First we recall semi-graphoid properties of conditional possibilistic independence, parameterized by a continuous $t$-norm, and find sufficient conditions for a class of Archimedean $t$-norms to have the graphoid property. Then we introduce Markov properties and factorization of possibility distributions (again parameterized by a continuous $t$-norm) and find the relationships between them. These results are accompanied by a number of counterexamples, which show that the assumptions of specific theorems are substantial.

## 1 INTRODUCTION

Conditional independence and Markov properties are fundamental notions of graphical modeling; therefore they are strongly connected with the application of probability theory to artificial intelligence. Complexity of practical problems that are of primary interest in the field of artificial intelligence usually results in the necessity to construct models with the aid of a great number of variables: more precisely, hundreds or thousands rather than tens. Processing distributions of such dimensionality would not be possible without some tools allowing us to reduce demands on computer memory. Conditional independence and Markov properties, which are among such tools, allow expression of these multidimensional distributions by means of low-dimensional ones, and therefore to substantially decrease demands on computer memory.

For three centuries, probability theory was the only mathematical tool at our disposal for uncertainty quantification and processing. As a result, many important theoretical and practical advances have been achieved in this field. However, during the last thirty years some new mathematical tools have emerged as alternatives to probability theory. They are used in situations whose nature of uncertainty does not meet the requirements of probability theory, or those in which probabilistic approaches employ criteria that are too strict. Nevertheless, probability theory has always served as a source of inspiration for the development of these nonprobabilistic calculi and these calculi have been continually confronted with probability theory and mathematical statistics from various points of view.

Good examples of this fact include the numerous papers studying conditional independence in various calculi (de Campos and Huete 1999) (Fonck 1994) (Shenoy 1994) (Spohn 1980). With this paper, we will continue our efforts in the area of possibility theory (Vejnarová 1999), attempting to unify the conditional independence notion. We will follow de Cooman's measure-theoretic approach (de Cooman 1997) to possibility theory, but for purposes of this paper we have chosen to forego generality in favour of simplicity.

## 2 BASIC TERMINOLOGY

In this section we will provide a brief overview of basic notions and results from (de Cooman 1997) and in Section 2.1 also from (de Baets et al. 1999) that are necessary for understanding the sections that follow.

### 2.1 TRIANGULAR NORMS

A *triangular norm* (or a *t-norm*) $T$ is a binary operator on $[0, 1]$ (i.e. $T : [0, 1]^2 \to [0, 1]$) satisfying the following three conditions:

(i) *boundary conditions:* for any $a \in [0,1]$

$$T(1,a) = a, \qquad T(0,a) = 0;$$

(ii) *isotonicity:* for any $a_1, a_2, b_1, b_2 \in [0,1]$ such that $a_1 \leq a_2, b_1 \leq b_2$

$$T(a_1, b_1) \leq T(a_2, b_2);$$

(iii) *associativity and commutativity:* for any $a, b, c \in [0,1]$

$$T(T(a,b),c) = T(a, T(b,c)),$$
$$T(a,b) = T(b,a).$$

A $t$-norm $T$ is called *continuous*, if $T$ is a continuous function. In this case commutativity is just a consequence of the remaining conditions.

There exist three distinct continuous $t$-norms, which will be studied in this paper:

(i) *Gödel's $t$-norm:*[1] $T_G(a,b) = \min(a,b)$;

(ii) *product $t$-norm:* $T_P(a,b) = a \cdot b$;

(iii) *Lukasziewicz' $t$-norm:* $T_L(a,b) = \max(0, a+b-1)$.

Because of its associativity, any $t$-norm $T$ can be extended to an $n$-ary operator $T^n : [0,1]^n \to [0,1]$, namely, in the following way

$$T(T^{n-1}(x_1, \ldots x_{n-1}), x_n) = T^n(x_1, \ldots, x_n).$$

A $t$-norm $T$ is called *Archimedean* iff, for any $(x,y) \in (0,1)^2$, there exists $n \in \mathbf{N}$ such that

$$T^n(x, \ldots, x) < y.$$

Let us note that both product and Lukasziewicz' $t$-norms are Archimedean, but Gödel's is not.

A $t$-norm $T$ is called *strict* iff it is continuous and, for any $x \in (0,1)$, $T(x, \cdot)$ is strictly increasing. A $t$-norm $T$ is called *nilpotent* iff it is continuous, Archimedean and not strict, i.e., any continuous Archimedean $t$-norm is either strict or nilpotent. $\varphi$-transform of $T$ is a $t$-norm $T_\varphi$ defined by

$$T_\varphi(x,y) = \varphi^{-1}(T(\varphi(x), \varphi(y))). \qquad (1)$$

It can easily be seen that product $t$-norm is strict and Lukasziewicz' $t$-norm is nilpotent. Moreover, the following proposition (see e.g. (de Baets et al. 1999)) holds true:

**Proposition 1** *(i) A $t$-norm $T$ is strict if and only if there exists a [0,1]-automorphism $\varphi$ such that $T$ is the $\varphi$-transform of the product $t$-norm.*

*(ii) A $t$-norm $T$ is nilpotent if and only if there exists a [0,1]-automorphism $\psi$ such that $T$ is the $\psi$-transform of Lukasziewicz' $t$-norm.*

---

[1] Gödel's $t$-norm is sometimes also called *minimum $t$-norm*. But we prefer the first name, since its residual (see the last paragraph of this subsection) coincides with Gödel's implication well-known from fuzzy logic.

Let $x,y \in [0,1]$ and $T$ be a $t$-norm. We will call an element $z \in [0,1]$ $T$-*inverse* of $x$ w.r.t. $y$ iff $T(z,x) = T(x,z) = y$. It need no be defined uniquely (if it exists) and this ambiguity can cause serious problems in some cases. Therefore $T$-*residual* $y \triangle_T x$ of $y$ by $x$, which is defined as

$$y \triangle_T x = \sup\{z \in [0,1] : T(z,x) \leq y\},$$

is often preferred. It is well-known (see e.g. (de Cooman 1997), (de Baets et al. 1999)) that for continuous $t$-norms $T$ the *maximal* $T$-inverse of $x$ w.r.t. $y$ equals to $y \triangle_T x$. Thus, for our famous $t$-norms we get

$$y \triangle_{T_G} x = \begin{cases} y & \text{if } x > y, \\ 1 & \text{otherwise,} \end{cases}$$

$$y \triangle_{T_P} x = \begin{cases} \frac{y}{x} & \text{if } 0 \leq y < x, \\ 1 & \text{otherwise,} \end{cases}$$

$$y \triangle_{T_L} x = \begin{cases} y - x + 1 & \text{if } x > y, \\ 1 & \text{otherwise.} \end{cases}$$

Let us note that if $T_\varphi$ is a $\varphi$-transform of $T$ then

$$x \triangle_{T_\varphi} y = \varphi^{-1}(\varphi(x) \triangle_T \varphi(y)). \qquad (2)$$

## 2.2 POSSIBILITY MEASURES AND DISTRIBUTIONS

Let $\mathbf{X}$ be a set called *universe of discourse*, which is supposed to contain at least two elements. A *possibility measure* $\Pi$ is a mapping from the power set $\mathcal{P}(\mathbf{X})$ of $\mathbf{X}$ to the real unit interval $[0,1]$ satisfying the following requirement: for any family $\{A_j, j \in J\}$ of elements of $\mathcal{P}(\mathbf{X})$

$$\Pi(\bigcup_{j \in J} A_j) = \sup_{j \in J} \Pi(A_j).$$

$\Pi$ is called *normal* iff $\Pi(\mathbf{X}) = 1$. Within this paper we will deal only with normal possibility measures.

For any $\Pi$ there exists a mapping $\pi : \mathbf{X} \to [0,1]$ such that for any $A \in \mathcal{P}(\mathbf{X})$, $\Pi(A) = \sup_{x \in A} \pi(x)$. $\pi$ is called a *distribution* of $\Pi$. This function is a possibilistic counterpart of a density function in probability theory. Let us note that (if $\mathbf{X}$ is finite) $\Pi$ is normal iff $\pi(x) = 1$ for at least one $x \in \mathbf{X}$.

Now, let us consider an arbitrary possibility measure $\Pi$ defined on a product universe of discourse $\mathbf{X} \times \mathbf{Y}$. The *marginal possibility measure* is then defined by the equality

$$\Pi_X(A) = \Pi(A \times \mathbf{Y})$$

for any $A \subset \mathbf{X}$ and the *marginal possibility distribution* by the corresponding expression

$$\pi_X(x) = \sup_{y \in \mathbf{Y}} \pi(x,y) \qquad (3)$$

for any $x \in \mathbf{X}$. In what follows, we will omit the subscript if there are no doubts which marginal we have in mind.

### 2.2.1 Almost everywhere equality

A mapping $h : \mathbf{X} \to [0,1]$ is called *fuzzy variable*. The set of fuzzy variables on $\mathbf{X}$ will be denoted by $\mathcal{G}(\mathbf{X})$.

Let $T$ be a $t$-norm on $[0,1]$. For any possibility measure $\Pi$ on $\mathbf{X}$ with distribution $\pi$, we define the following binary relation on $\mathcal{G}(\mathbf{X})$. For $h_1$ and $h_2$ in $\mathcal{G}(\mathbf{X})$ we say that $h_1$ and $h_2$ are $(\Pi, T)$-*equal almost everywhere* (and write $h_1 \stackrel{(\Pi,T)}{=} h_2$) iff for any $x \in X$

$$T(h_1(x), \pi(x)) = T(h_2(x), \pi(x)).$$

The strength of this equality is dependent not only on the possibility measure (as in probabilistic framework) but also on the choice of $t$-norm. For example, for product $t$-norm it means that the equivalent functions may differ only on a set $E$ such that $\Pi(E) = 0$, while for Gödel's one the classes of equivalence are much wider. For more details see (de Cooman 1997).

This notion is very important for the definition of conditional possibility distribution as well as for that of (conditional) independence.

### 2.2.2 Conditioning

*Conditional possibility distribution* is defined as *any* solution of the equation

$$\pi_{XY}(x,y) = T(\pi_Y(y), \pi_{X|Y}(x|y)), \quad (4)$$

for any $(x,y) \in \mathbf{X} \times \mathbf{Y}$. The solution of this equation is not unique (in general), but the ambiguity vanishes when almost everywhere equality is considered. We are able to obtain a representative of these conditional possibility distributions (if $T$ is a continuous $t$-norm) by taking the residual

$$\pi_{X|Y}(x|\cdot) \stackrel{(\Pi_Y,T)}{=} \pi_{XY}(x,\cdot) \triangle_T \pi_Y(\cdot), \quad (5)$$

which is defined as the greatest solution of the equation (4) (cf. also Section 2.1).

We note that if we use product $t$-norm, we will obtain Dempster's rule of conditioning (Dempster 1967), Lukasziewicz' $t$-norm corresponds to "Lukasziewicz'" rule of conditioning[2] (Fonck 1994), Gödel's $t$-norm leads to Hisdal's rule of conditioning (Hisdal 1978), and the choice of Gödel's $t$-norm together with (5) gives the modification of Hisdal's rule proposed by (Dubois and Prade 1988).

### 2.2.3 Independence

De Cooman considered two variables $X$ and $Y$ *possibilistically $T$-independent* iff for any $F_X \in X^{-1}(\mathcal{P}(\mathbf{X}))$, $F_Y \in Y^{-1}(\mathcal{P}(\mathbf{Y}))$ and any $G_X \in \{F_X, F_X^C\}, G_Y \in \{F_Y, F_Y^C\}$

$$\Pi(G_X \cap G_Y) = T(\Pi(G_X), \Pi(G_Y)).$$

---

[2] This conditioning rule is somewhat questionable: from impossible combination of events we can obtain that one of them conditioned by the other is "somewhat possible"; for more discussion see (Vejnarová 1999).

From this definition it immediately follows that the independence notion is parameterized by $T$. This fact was not mentioned in Zadeh's and Hisdal's works since they used only one $t$-norm, Gödel's $t$-norm. Furthermore, de Cooman's definition reveals the relationship between independence of variables and events; for more details see (de Cooman 1997).

What is more important, from the viewpoint taken in this paper, is the following theorem which is an immediate consequence of Proposition 2.6. in (de Cooman 1997).

**Theorem 1** *Let us assume that $t$-norm $T$ is continuous. Then the following propositions are equivalent.*

*(i)* $X$ *and* $Y$ *are $T$-independent.*

*(ii) For any $x \in \mathbf{X}$ and $y \in \mathbf{Y}$*

$$\pi_{XY}(x,y) = T(\pi_X(x), \pi_Y(y)).$$

*(iii) For any $x \in \mathbf{X}$ and $y \in \mathbf{Y}$*

$$\begin{aligned}T(\pi_X(x), \pi_Y(y)) &= T(\pi_{X|Y}(x|y), \pi_Y(y)) = \\ &= T(\pi_{Y|X}(y|x), \pi_X(x)).\end{aligned}$$

This theorem shows that the notion of independence defined by de Cooman is equivalent (for $T = \min$) to Zadeh's notion of noninteractivity (Zadeh 1978) and, in a sense, also to Hisdal's notion of independence (Hisdal 1978) — if the equality sign in her definition is substituted by almost everywhere equality.

## 3 CONDITIONAL INDEPENDENCE

Among the properties satisfied by the ternary relation $I(X,Y|Z)$ of independence, regardless the framework in question, the following are of principal importance:

(A1) $I(X,Y|Z) \to I(Y,X|Z)$          symmetry,

(A2) $I(X,YZ|W) \to I(X,Z|W)$      decomposition,

(A3) $I(X,YZ|W) \to I(X,Y|ZW)$       weak union,

(A4) $[I(X,Y|ZW) \wedge I(X,Z|W)] \to I(X,YZ|W)$
                                              contraction,

(A5) $[I(X,Y|ZW) \wedge I(X,Z|YW)] \to I(X,YZ|W)$
                                              intersection.

### 3.1 CONDITIONAL $T$-INDEPENDENCE

In light of the above-mentioned facts (cf. Theorem 1), in (Vejnarová 1998) we defined the conditional possibilistic independence in the following way. Variables $X$ and $Y$ are *possibilistically conditionally $T$-independent* given $Z$ $(I_T(X,Y|Z))$ iff for any pair $(x,y) \in \mathbf{X} \times \mathbf{Y}$

$$\pi_{XY|Z}(x,y|\cdot) \stackrel{(\Pi_Z,T)}{=} T(\pi_{X|Z}(x|\cdot), \pi_{Y|Z}(y|\cdot)). \quad (6)$$

Let us stress again that we do not deal with pointwise equality, but with *almost everywhere equality* in contrast to the conditional noninteractivity (Fonck, 1994). The following theorem proven in (Vejnarová, 1999) is a "conditional counterpart" of Theorem 1.

**Theorem 2** *Let us assume that t-norm $T$ is continuous. Then the following propositions are equivalent.*

*(i) $X$ and $Y$ are $T$-independent given $Z$.*

*(ii) For any $x \in \mathbf{X}$, $y \in \mathbf{Y}$ and $z \in \mathbf{Z}$*
$$T(\pi_{X|YZ}(x|y,z), \pi_{YZ}(y,z)) = \\ = T(\pi_{X|Z}(x|z), \pi_{YZ}(y,z)).$$

Theorem 2 unifies the notions of conditional noninteractivity (Fonck 1994) and various notions of conditional independence, which can be found in (de Campos and Huete 1999), (Fonck 1994) (if we use specific $t$-norms) in the sense that *pointwise* equalities are substituted by *almost everywhere* ones.

It should also be mentioned that one particular type of conditional independence $I_T(X,Y|Z)$ has been proposed in (de Campos and Huete 1995) for Gödel's $t$-norm.

### 3.2 SEMI-GRAPHOID PROPERTIES OF CONDITIONAL $T$-INDEPENDENCE

In (Vejnarová 1999) we studied semi-graphoid properties of the relation $I_T(X,Y|Z)$ and proved

**Theorem 3** *For any continuous t-norm $T$, the relation $I_T(X,Y|Z)$ satisfies (A1) – (A4).*

Property (A5) is not fulfilled in general, which is obvious from Example 1. This example was previously published in (Vejnarová 1999), but it seems useful to have it reiterated here, as we will see later.

**Example 1** Let $\mathbf{X} = \mathbf{Y} = \mathbf{Z} = \{0,1\}$ and
$$\pi_{XYZ}(x,y,z) = \begin{cases} 1 & \text{if } x = y = z, \\ 0 & \text{else.} \end{cases}$$
Then
$$\pi_{XY}(x,y) = \begin{cases} 1 & \text{if } x = y, \\ 0 & \text{else,} \end{cases}$$
$$\pi_{XZ}(x,z) = \begin{cases} 1 & \text{if } x = z, \\ 0 & \text{else,} \end{cases}$$
$$\pi_{YZ}(y,z) = \begin{cases} 1 & \text{if } y = z, \\ 0 & \text{else,} \end{cases}$$
and
$$\pi_Y \equiv \pi_Z \equiv 1.$$
Then, for *any* continuous $t$-norm,[3]
$$\pi_{XY|Z}(x,y|z) = T(\pi_{X|Z}(x|z), \pi_{Y|Z}(y|z)),$$
$$\pi_{XZ|Y}(x,z|y) = T(\pi_{X|Y}(x|y), \pi_{Z|Y}(z|y)),$$

---
[3]Let us note that the following equalities are pointwise, since $\pi_Y \equiv \pi_Z \equiv 1$.

for any $(x,y,z) \in \mathbf{X} \times \mathbf{Y} \times \mathbf{Z}$, but e.g.
$$\pi_{XYZ}(1,0,0) \neq T(\pi_X(1), \pi_{YZ}(0,0)),$$
i.e., $I_T(X,Y|Z)$ and $I_T(X,Z|Y)$ hold, but $I_T(X,YZ|\emptyset)$ does not. $\diamond$

Therefore we concluded:

**Proposition 2** *There exists no t-norm $T$ such that $I_T(X,Y|Z)$ satisfies (A1)–(A5) for arbitrary possibility distribution.*

### 3.3 GRAPHOID PROPERTY FOR ARCHIMEDEAN $T$-NORMS

This fact perfectly corresponds to the properties of probabilistic conditional independence. In probability theory, (A5) need not be satisfied if the probability distribution is not strictly positive. In this case the conditional probability distributions need not be defined uniquely. In possibility theory this non-uniqueness is caused by the use of $t$-norms. If we adopt the axiomatic approach presented in this paper, satisfaction of (A5) depends on the choice of a $t$-norm and on the properties of the possibility distribution in question. If we choose an Archimedean $t$-norm, (A5) is always satisfied by strictly positive possibility distributions, as expressed by Theorem 4.

**Lemma 1** *Let $T$ be a strict t-norm and $\pi(x,y,z)$ be strictly positive. Then the following statements are equivalent.*

*(i) Variables $X$ and $Y$ are conditionally $T$-independent given $Z$.*

*(ii) Joint distribution of $X, Y$ and $Z$ has a form*
$$\pi(x,y,z) = \varphi^{-1}(\rho_1(x,z) \cdot \rho_2(y,z)),$$
*for some $\rho_1$ and $\rho_2$, such that $\rho_1(x,z) \cdot \rho_2(y,z) \in [0,1]$ for all $(x,y,z) \in \mathbf{X} \times \mathbf{Y} \times \mathbf{Z}$, and $[0,1]$-automorphism $\varphi$, such that $T$ is $\varphi$-transform of product t-norm.*

*Proof.* Let (i) be satisfied. Then
$$\begin{aligned}\pi(x,y,z) &= T(T(\pi(x|z), \pi(y|z)), \pi(z)) = \\ &= T(\pi(x|z), \pi(y,z)) = \\ &= \varphi^{-1}(\varphi(\pi(x|z)) \cdot \varphi(\pi(y,z)))\end{aligned}$$
by (1) and (i) in Proposition 1, and therefore (ii) is obviously fulfilled (e.g. $\rho_1(x,z) = \varphi(\pi(x|z))$ and $\rho_2(y,z) = \varphi(\pi(y,z))$).

Let (ii) be satisfied. Then by (2), (1) and (i) in Proposition 1
$$\begin{aligned}\pi_{XY|Z}&(x,y,z) = \\ &= \varphi^{-1}(\rho_1(x,z) \cdot \rho_2(y,z)) \triangle_T \varphi^{-1}(\rho_1(z) \cdot \rho_2(z)) = \\ &= \varphi^{-1}\left(\frac{\rho_1(x,z) \cdot \rho_2(y,z)}{\rho_1(z) \cdot \rho_2(z)}\right) =\end{aligned}$$

$$= \varphi^{-1} \left( \frac{\rho_1(x,z) \cdot \rho_2(z)}{\rho_1(z) \cdot \rho_2(z)} \cdot \frac{\rho_1(z) \cdot \rho_2(y,z)}{\rho_1(z) \cdot \rho_2(z)} \right) =$$

$$= \varphi^{-1} \left( \frac{\varphi(\varphi^{-1}(\rho_1(x,z) \cdot \rho_2(z)))}{\varphi(\varphi^{-1}(\rho_1(z) \cdot \rho_2(z)))} \cdot \right.$$
$$\left. \cdot \frac{\varphi(\varphi^{-1}(\rho_1(z) \cdot \rho_2(y,z)))}{\varphi(\varphi^{-1}(\rho_1(z) \cdot \rho_2(z)))} \right) =$$

$$= \varphi^{-1} \left( \frac{\varphi(\pi_{XZ}(x,z))}{\varphi(\pi_Z(z))} \cdot \frac{\varphi(\pi_{YZ}(y,z))}{\varphi(\pi_Z(z))} \right) =$$

$$= \varphi^{-1} \left( \varphi(\pi_{X|Z}(x|z)) \cdot \varphi(\pi_{Y|Z}(y|z)) \right)$$

$$= T(\pi_{X|Z}(x|z), \pi_{Y|Z}(y|z)),$$

i.e. (i) is satisfied. □

**Lemma 2** *Let $T$ be a nilpotent t-norm and $\pi(x, y, z)$ be strictly positive. Then the following statements are equivalent.*

*(i) Variables $X$ and $Y$ are conditionally $T$-independent given $Z$.*

*(ii) Joint distribution of $X, Y$ and $Z$ has a form*
$$\pi(x,y,z) = \psi^{-1}(\rho_1(x,z) + \rho_2(y,z)),$$
*for some $\rho_1$ and $\rho_2$, such that $\rho_1(x,z) + \rho_2(y,z) \in [0,1]$ for all $(x,y,z) \in \mathbf{X} \times \mathbf{Y} \times \mathbf{Z}$, and $[0,1]$-automorphism $\psi$, such that $T$ is $\psi$-transform of Lukasziewicz t-norm.*

*Proof.* Let (i) be satisfied. Then we have (analogous to the proof of Lemma 1)
$$\pi(x,y,z) = T(T(\pi(x|z), \pi(y|z)), \pi(z)) =$$
$$= \psi^{-1}(\psi(\pi(x|z)) + \psi(\pi(y,z)) - 1)$$
by (1) and (ii) in Proposition 1, and therefore (ii) is obviously satisfied (e.g. $\rho_1(x,z) = \psi(\pi(x|z))$ and $\rho_2(y,z) = \psi(\pi(y,z)) - 1$).

Let (ii) be satisfied. Then by (2), (1) and (ii) in Proposition 1
$$\pi_{XY|Z}(x,y,z) =$$
$$= \psi^{-1}(\rho_1(x,z) + \rho_2(y,z)) \triangle_T$$
$$\triangle_T \psi^{-1}(\rho_1(z) + \rho_2(z)) =$$
$$= \psi^{-1}((\rho_1(x,z) + \rho_2(y,z)) -$$
$$- (\rho_1(z) + \rho_2(z)) + 1) =$$
$$= \psi^{-1}((\rho_1(x,z) + \rho_2(z)) - (\rho_1(z) + \rho_2(z)) +$$
$$+ (\rho_1(z) + \rho_2(y,z)) - (\rho_1(z) + \rho_2(z)) + 1) =$$
$$= \psi^{-1}((\psi(\psi^{-1}(\rho_1(x,z) + \rho_2(z))) -$$
$$- \psi(\psi^{-1}(\rho_1(z) + \rho_2(z))) +$$
$$+ \psi(\psi^{-1}(\rho_1(z) + \rho_2(y,z))) -$$
$$- \psi(\psi^{-1}(\rho_1(z) + \rho_2(z))) + 1) =$$
$$= \psi^{-1}(\psi(\pi_{XZ}(x,z)) - \psi(\pi_Z(z)) +$$
$$+ \psi(\pi_{YZ}(y,z)) - \psi(\pi_Z(z)) + 1) =$$
$$= \psi^{-1}(\psi(\pi_{X|Z}(x|z)) - 1 + \psi(\pi_{Y|Z}(y|z))) =$$
$$= T(\pi_{X|Z}(x|z), \pi_{Y|Z}(y|z)),$$
i.e. (i) is satisfied. □

**Theorem 4** *Let $T$ be an Archimedean t-norm and $\pi$ be strictly positive possibility distribution. Then (A5) is also satisfied.*

*Proof.* Let $I_T(X,Y|ZW)$ and $I_T(X,Z|YW)$ be satisfied. Due to the fact that t-norm $T$ is Archimedean, $\pi$ has either the form

$$\pi_{XYZW}(x,y,z,w) =$$
$$= \varphi^{-1}(\rho_1(x,z,w) \cdot \rho_2(y,z,w)) =$$
$$= \varphi^{-1}(\sigma_1(x,y,w) \cdot \sigma_2(y,z,w))$$

due to Lemma 1, or the form

$$\pi_{XYZW}(x,y,z,w) =$$
$$= \psi^{-1}(\rho_1(x,z,w) + \rho_2(y,z,w)) =$$
$$= \psi^{-1}(\sigma_1(x,y,w) + \sigma_2(y,z,w))$$

due to Lemma 2.

Thus, we have, in the first case, for all $z$
$$\sigma_1(x,y,w) = \frac{\rho_1(x,z,w) \cdot \rho_2(y,z,w)}{\sigma_2(y,z,w)}.$$

Choosing a fixed $z = z_0$ we have
$$\sigma_1(x,y,w) = f(x,w) \cdot g(y,w)$$

where
$$f(x,w) = \rho_1(x, z_0, w)$$

and
$$g(y,w) = \frac{\rho_2(y, z_0, w)}{\sigma_2(y, z_0, w)}.$$

Therefore
$$\pi_{XYZW}(x,y,z,w) =$$
$$= \varphi^{-1}(f(x,w) \cdot g(y,w) \cdot \sigma_2(y,z,w))$$

and $f(x,w) \cdot g(y,w) \cdot \sigma_2(y,z,w) \in [0,1]$ for all $(x,y,z,w) \in \mathbf{X} \times \mathbf{Y} \times \mathbf{Z} \times \mathbf{W}$, since both $\rho_1(x,z,w) \cdot \rho_2(y,z,w) \in [0,1]$ and $\sigma_1(x,y,w) \cdot \sigma_2(y,z,w) \in [0,1]$ for all $(x,y,z,w) \in \mathbf{X} \times \mathbf{Y} \times \mathbf{Z} \times \mathbf{W}$, and hence $I_T(X, YZ|W)$ (again due to Lemma 1) as desired.

The proof in the second case is completely analogous. We have that for all $z$
$$\sigma_1(x,y,w) = \rho_1(x,z,w) + \rho_2(y,z,w) - \sigma_2(y,z,w).$$

Choosing a fixed $z = z_0$ we have
$$\sigma_1(x,y,w) = f(x,w) + g(y,w)$$

where
$$f(x,w) = \rho_1(x, z_0, w)$$

and
$$g(y,w) = \rho_2(y, z_0, w) - \sigma_2(y, z_0, w)$$

Therefore
$$\pi_{XYZW}(x,y,z,w) =$$
$$= \psi^{-1}(f(x,w) + g(y,w) + \sigma_2(y,z,w))$$

and $f(x,w) + g(y,w) + \sigma_2(y,z,w) \in [0,1]$ for all $(x,y,z,w) \in \mathbf{X} \times \mathbf{Y} \times \mathbf{Z} \times \mathbf{W}$, (the arguments are

the same as above), and hence $I_T(X, YZ|W)$ (again due to Lemma 2) as desired. □

Since Gödel's $t$-norm is not Archimedean, Theorem 4 does not hold true for it, as can be seen from the following example:

**Example 2** Let $\mathbf{X} = \mathbf{Y} = \mathbf{Z} = \{0, 1\}$ and

$$\pi_{XYZ}(x, y, z) = \begin{cases} 1 & \text{if } x = y = z, \\ \frac{1}{2} & \text{else.} \end{cases}$$

Then analogous to Example 1

$$\begin{aligned} \pi_{XY|Z}(x,y|z) &= \min(\pi_{X|Z}(x|z), \pi_{Y|Z}(y|z)), \\ \pi_{XZ|Y}(x,z|y) &= \min(\pi_{X|Y}(x|y), \pi_{Z|Y}(z|y)), \end{aligned}$$

for any $(x, y, z) \in \mathbf{X} \times \mathbf{Y} \times \mathbf{Z}$, but e.g.

$$\pi_{XYZ}(1, 0, 0) \neq \min \pi_X(1), \pi_{YZ}(0, 0)),$$

i.e. $I_G(X, Y|Z)$ and $I_G(X, Z|Y)$ hold, but $I_G(X, YZ|\emptyset)$ does not. ◇

## 4 MARKOV PROPERTIES AND FACTORIZATION

Before introduction of Markov properties[4] and factorization, let us present a few necessary notions from graph theory.

A *graph* is a pair $G = (V, E)$, where $V$ is a finite set of *vertices* and the set of *edges* $E$ is a subset of the set $V \times V$ of (unordered) pairs of distinct vertices. A subset of the vertex set $A \subseteq V$ *induces* a subgraph $G_A = (A, E_A)$, where the edge set $E_A = E \cap (A \times A)$ is obtained from $G$ by keeping edges with both end points in $A$.

A graph is *complete* if all vertices are joined by a line. A subset is complete if it induces a complete subgraph. A maximal (with respect to set inclusion) complete subset is called *clique*.

If there is a line between $a \in V$ and $b \in V$, $a$ and $b$ are said to be *adjacent*, otherwise they are *non-adjacent*. The *boundary* $bd(A)$ of a subset $A$ of vertices is the set of vertices in $V \setminus A$ that are adjacent to vertices in $A$. The *closure* of $A$ is $cl(A) = A \cup bd(A)$.

If there is a *path* from $a$ to $b$ (a sequence $a = a_0, a_1, \ldots, a_n = b$ of distinct vertices such that $(a_{i-1}, a_i) \in E$ for all $i = 1, \ldots, n$) we say that $a$ and $b$ are in the same *connectivity component*. A subset $S \subseteq V$ is called an $(a, b)$-*separator* if all paths from $a$ to $b$ intersect $S$. The subset $C$ *separates* $A$ from $B$ if it is an $(a, b)$-separator for every $a \in A$ and $b \in B$.

---

[4]Markov properties studied in this section are, in a sense, generalization of the well-known property defining Markov chains; see e.g. (Meyer 1967)

### 4.1 MARKOV PROPERTIES

Now, let us consider the conditional independence in a special situation: we have a graph $G = (V, E)$ and a finite collection of variables $(X_i)_{i \in V}$ taking their values in $(\mathbf{X}_i)_{i \in V}$. For $A \subset V$ we define

$$\mathbf{X}_A = \times_{i \in A} \mathbf{X}_i$$

and $\mathbf{X} = \mathbf{X}_V$. Here we will use, for simplicity sake, the symbol $I_T(A, B|C)$ instead of $I_T(X_A, X_B|X_C)$.

With any undirected graph $G = (V, E)$ and a collection of variables $(X_i)_{i \in V}$ we can (analogous to probability theory) associate three different Markov properties. A possibility measure is said to obey

(P) *the pairwise Markov property*, relative to $G$ and $T$ if, for any pair $(i, j)$ of non-adjacent vertices,

$$I_T(i, j | V \setminus \{i, j\});$$

(L) *the local Markov property*, relative to $G$ and $T$ if, for any vertex $i \in V$,

$$I_T(i, V \setminus cl(i) | bd(i));$$

(G) *the global Markov property*, relative to $G$ and $T$ if, for any triple $(A, B, S)$ of disjoint subsets of $V$ such that $S$ separates $A$ from $B$ in $G$,

$$I_T(A, B|S).$$

The global Markov property (G) gives the general criterion for deciding whether two groups of variables $A$ and $B$ are conditionally independent, given a third group of variables $S$. It is the strongest property, as can be seen from Theorem 5.

The following two theorems (Theorem 5 and Theorem 6) are presented without proofs. Their proofs can be found in (Lauritzen 1996), since they completely depend on semi-graphoid and graphoid properties, respectively, and not on the distributions in question.

**Theorem 5** *For any undirected graph $G$ and any possibility distribution on $\mathbf{X}$ it holds true that*

$$(G) \Longrightarrow (L) \Longrightarrow (P). \tag{7}$$

These three properties are really different (analogous to probabilistic framework), as can be seen from the following two examples, inspired by Example 3.6 and Example 3.5 from (Lauritzen 1996).

**Example 3** Let $\mathbf{X} = \mathbf{Y} = \mathbf{Z} = \{0, 1\}$ and the joint distribution of three bivariate variables $X, Y$ and $Z$ on the graph

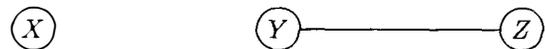

be defined as in Example 1. From the results obtained in that example, we can see that this possibility distribution satisfies the pairwise Markov property (P), since $I_T(X, Z|Y)$, but does not satisfy the local Markov property (L), since not $I_T(X, YZ|\emptyset)$. ◇

**Example 4** Let $\mathbf{U} = \mathbf{W} = \mathbf{X} = \mathbf{Y} = \mathbf{Z} = \{0,1\}$ and the joint possibility distribution of five bivariate variables be defined on the graph $G$

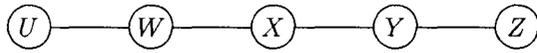

as follows

$$\pi(u,w,x,y,z) = \begin{cases} 1 & \begin{array}{l}\text{if } u=w=1 \& x=y=z=0, \\ \text{or } u=w=x=0 \& y=z=1,\end{array} \\ 0 & \text{else.} \end{cases}$$

It is easy (but somewhat time-consuming) to verify that this distribution satisfies (L) (for any continuous $t$-norm $T$) but, since

$$T(\pi_{UW|X}(0,0|0), \pi_{YZ|X}(0,0|0)) = 1 \neq$$
$$\neq 0 = \pi_{UWYZ|X}(0,0,0,0|0),$$

it does not satisfy (G) (again for arbitrary continuous $t$-norm $T$). $\diamond$

Note that it was necessary to use five variables in Example 4, since $(G) \iff (L)$ up to $|V| = 4$. To see this, let us consider three disjoint subsets of $V$: $A, B, S$, such that $S$ separates $A$ from $B$. Let us assume that each of these subsets is nonempty (otherwise the equivalence is trivial). Then at least one of $A$ and $B$ is a singleton $\{i\}$, and therefore $bd\{i\} \subseteq S$. Then (G) follows from (L) using weak union (A3).

**Theorem 6** *If a possibility distribution on $\mathbf{X}$ is such that (A5) holds true for disjoint subsets $A, B, C, D$ substituted for $X, Y, Z, W$ then*

$$(G) \iff (L) \iff (P).$$

Due to the results obtained in the previous section (Theorem 4), we immediately get the following

**Corollary 1** *Let $T$ be an Archimedean $t$-norm and $\pi$ a strictly positive possibility distribution. Then*

$$(G) \iff (L) \iff (P).$$

### 4.2 FACTORIZATION

Factorization is a property even more strict than the global Markov property (G). We will say that a possibility distribution $\pi$ *factorizes* (has a property (F)) with respect to $G$ and $T$, if, for all complete subsets $A \subseteq V$, there exist non-negative functions $\psi_A$ of $x_A$ such that $\pi$ has the form

$$\pi(x) = T^{|\mathcal{A}|}(\psi_{A_1}(x_{A_1}), \ldots, \psi_{A_{|\mathcal{A}|}}(x_{A_{|\mathcal{A}|}})),$$

where $\mathcal{A}$ denotes the set of all complete subsets of $V$. The functions $\psi_A$ are not uniquely determined (in general), since they can be "multiplied" in several ways.

Without loss of generality we can assume (cf. (Lauritzen 1996)) that only cliques (maximum complete subgraphs) appear as the sets, i. e., that

$$\pi(x) = T^{|\mathcal{C}|}(\psi_{C_1}(x_{C_1}), \ldots, \psi_{C_{|\mathcal{C}|}}(x_{C_{|\mathcal{C}|}})), \quad (8)$$

where $\mathcal{C}$ denotes the set of all cliques of $G$.

**Theorem 7** *Let $T$ be an Archimedean $t$-norm. Then, for any undirected graph $G$ and any possibility distribution $\pi$ on $\mathbf{X}$,*

$$(F) \Longrightarrow (G).$$

*Proof.* Let $A, B, S$ be three disjoint subsets of $V$ such that $S$ separates $A$ from $B$. Let $\bar{A}$ denote the connectivity component in subgraph of $G$ induced by $V \setminus S$ containing $A$ and let $\bar{B} = V \setminus (\bar{A} \cup S)$. Since $A$ and $B$ are separated by $S$, their elements must be in different connectivity components of the subgraph induced by $V \setminus S$, and any clique of $G$ is a subset of either $\bar{A} \cup S$ or $\bar{B} \cup S$. Let us denote by $\mathcal{C}_A$ the set of cliques contained in $\bar{A} \cup S$. Using (8), we get

$$\pi(x) = T^{|\mathcal{C}|}(\psi_{C_1}(x_{C_1}), \ldots, \psi_{C_{|\mathcal{C}|}}(x_{C_{|\mathcal{C}|}})) =$$
$$= T\Big(T^{|\mathcal{C}_A|}(\psi_{C_1}(x_{C_1}), \ldots, \psi_{C_{|\mathcal{C}_A|}}(x_{C_{|\mathcal{C}_A|}})),$$
$$T^{|\mathcal{C}\setminus\mathcal{C}_A|}(\psi_{C'_1}(x_{C'_1}), \ldots, \psi_{C'_{|\mathcal{C}\setminus\mathcal{C}_A|}}(x_{C'_{|\mathcal{C}\setminus\mathcal{C}_A|}}))\Big) =$$
$$= T(\rho_1(x_{\bar{A}\cup S}), (\rho_2(x_{\bar{B}\cup S})).$$

Since $T$ is Archimedean, we will obtain, due to Proposition 1 and Lemma 1 or Lemma 2, that $I_T(\bar{A}, \bar{B}|S)$. Application of decomposition (A2) gives $I_T(A, B|S)$ and the second one (due to symmetry (A1)) $I_T(A, B|S)$ as desired. □

The reverse of this implication is not valid, as can be seen from the following example:

**Example 5** Let $X, Y, Z$ and $W$ be four bivariate distributions defined on the graph with the joint possibility distribution $\pi$ defined as follows: any of the following combinations has possibility equal to 1, the remaining combinations are equal to 0:

$$(0,0,0,0) \quad (0,0,0,1) \quad (0,0,1,1) \quad (0,1,1,1)$$
$$(1,0,0,0) \quad (1,1,0,0) \quad (1,1,1,0) \quad (1,1,1,1).$$

From this definition we immediately see that $\pi_{X,Z} \equiv \pi_{Y,W} \equiv 1$ and therefore the conditional distributions equal unconditional ones (for any continuous $t$-norm). Therefore we have

$$\pi_{XZ|YW}(0,0|0,0) = 1 = T(\pi_{X|YW}(0|0,0), \pi_{Z|YW}(0|0,0))$$
$$\pi_{XZ|YW}(1,0|0,0) = 1 = T(\pi_{X|YW}(1|0,0), \pi_{Z|YW}(0|0,0))$$
$$\pi_{XZ|YW}(0,1|0,0) = 0 = T(\pi_{X|YW}(0|0,0), \pi_{Z|YW}(1|0,0))$$
$$\pi_{XZ|YW}(1,1|0,0) = 0 = T(\pi_{X|YW}(1|0,0), \pi_{Z|YW}(1|0,0))$$

and similarly for the other values of $Y$ and $W$. We will also get similar results for $\pi_{YW|XZ}$; therefore, we can see that the distribution $\pi$ satisfies the global Markov property (G) with respect any continuous $t$ norm $T$ and the chordless 4-cycle:

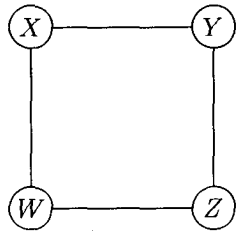

But $\pi$ does not factorize with respect to this graph. To see it, let us assume that $\pi$ factorizes. Then

$1 = \pi(0,0,0,0) =$
$= T^4(\phi_{XY}(0,0), \phi_{YZ}(0,0), \phi_{ZW}(0,0), \phi_{WX}(0,0)),$

but also

$0 = \pi(0,1,0,0) =$
$= T^4(\phi_{XY}(0,1), \phi_{YZ}(1,0), \phi_{ZW}(0,0), \phi_{WX}(0,0)).$

From this we have

$$T(\phi_{XY}(0,1), \phi_{YZ}(1,0)) = 0.$$

Since

$1 = \pi(1,1,0,0) =$
$= T^4(\phi_{XY}(1,1), \phi_{YZ}(1,0), \phi_{ZW}(0,0), \phi_{WX}(0,1)),$

it is evident that $\phi_{YZ}(1,0) = 1$, therefore $\phi_{XY}(0,1) = 0$. But then

$1 = \pi(0,1,1,1) \neq$
$\neq T^4(\phi_{XY}(0,1), \phi_{YZ}(1,1), \phi_{ZW}(1,1), \phi_{WX}(1,0)) = 0;$

therefore, it is evident that $\pi$ cannot factorize. $\diamond$

## 5 CONCLUSIONS

The goal of this paper was to promote de Cooman's measure-theoretic approach to possibility theory by demonstrating the parallels between the probabilistic and possibilistic approaches. We have shown that for the wide class of Archimedean $t$-norms, the conditional $T$-independence has the same properties as the conditional independence in probability theory. We also introduced Markov properties of possibility measures and demonstrated that the relationships between them are completely analogous with those of probability measures. And finally, after introducing the $T$-factorization, we proved that it implies the global Markov property for Archimedean $t$-norms.

There are still some open problems, which should be solved in the near future. The first one, perhaps most important, concerns Gödel's $t$-norm. Gödel's $t$-norm is the "classical" one in possibility theory, but many theorems presented in this paper do not hold true for it, since it is not Archimedean. Hence, the task is to find analogous theorems for Gödel's $t$-norm. The second one is to find an analogy to the Clifford-Hammersley theorem for possibility measures, i.e., to find conditions under which

$$(F) \iff (G) \iff (L) \iff (P).$$

The usefulness of such a theorem is obvious, since it would enable to us to determine whether or not a possibility distribution factorizes with respect to a given graph, simply by checking the pairwise Markov property.

## Acknowledgement

The research was financially supported by the grants MŠMT no. VS96008, GA ČR no. 201/98/1487 and KONTAKT ME 200/1998.